\documentclass{article}
\usepackage{spconf,amsmath,graphicx}
\usepackage{subfigure}
\usepackage{booktabs}
\usepackage{multirow}
\usepackage{soul, color}

\ninept

\title{Improving N-gram Language Models with Pre-Trained Deep Transformer}
\name{\begin{tabular}{c}Yiren Wang$^{1,}$\sthanks{This work is conducted during internship at Facebook.}, Hongzhao Huang$^{2}$, Zhe Liu$^{2}$, Yutong Pang$^{2}$, Yongqiang Wang$^{2}$, \\ ChengXiang Zhai$^{1}$, Fuchun Peng$^{2}$\end{tabular}}
\address{$^{1}$ University of Illinois at Urbana-Champaign, $^{2}$ Facebook AI}

\begin{document}
%
\maketitle
\begin{abstract}
Although $n$-gram language models (LMs) have been outperformed by the state-of-the-art neural LMs, they are still widely used in speech recognition due to its high efficiency in inference. In this paper, we demonstrate that $n$-gram LM can be improved by neural LMs through a text generation based data augmentation method. In contrast to previous approaches, we employ a large-scale general domain pre-training followed by in-domain fine-tuning strategy to construct deep Transformer based neural LMs. Large amount of in-domain text data is generated with the well trained deep Transformer to construct new $n$-gram LMs, which are then interpolated with baseline $n$-gram systems. Empirical studies on different speech recognition tasks show that the proposed approach can effectively improve recognition accuracy. In particular, our proposed approach brings significant relative word error rate reduction up to $6.0\%$ for domains with limited in-domain data. 
\end{abstract}


\section{Introduction}
\label{sec:intro}

$N$-gram language models (LMs) are widely used in the automatic speech recognition (ASR) systems due to its simplicity and high efficiency in inference. However, $n$-gram LMs suffer from performance bottleneck caused by the poor generalization to unseen $n$-grams and lack of ability to capture long range dependencies. Neural language models~\cite{bengio2003neural,mikolov2010recurrent,sundermeyer2012lstm} have overcome such deficiencies with distributed representation learning in the continuous space and achieved the state-of-the-art language modeling performances. However, the high computational cost hampers inference latency and makes it hard to directly integrate neural LMs into the first-pass decoding of an ASR system. Instead, neural LMs are commonly used in second-pass rescoring and have shown effectiveness in improving recognition accuracy~\cite{deoras2011fast,chan2016listen,xiong2018microsoft}. Still, since the N-best list or lattice for rescoring generally depends on $n$-gram LMs from the first-pass decoding, improving the performance of $n$-gram LMs is of great importance.

Different approaches have been proposed to improve $n$-gram LMs with neural LMs, including probability based methods that directly convert probabilities of neural LMs to the $n$-gram LMs~\cite{arisoy2013converting,wang2013converting,adel2014comparing}, and text generation based methods that leverage shallow recurrent neural networks to generate text for $n$-gram training~\cite{suzuki2019improvements}. Empirical studies have shown that the latter generally leads to better performances~\cite{adel2014comparing}. However, previous work in this line has not fully leveraged the state-of-the-art deep neural networks such as deep Transformer~\cite{vaswani2017attention,radford2018improving}, and is less applicable to situations where the in-domain data are too limited to train a good neural LM.

From another perspective, constructing good LMs depends on adequate high-quality training data. Unfortunately, in many cases, only limited in-domain data are accessible, making the data sparsity problem even more severe for $n$-gram LMs, and also introducing optimization difficulties for training neural LMs. Effectively leveraging the rich general domain corpora could help ease challenge and construct high-capacity neural networks with good generalization.

In this paper, we propose a new text generation based data augmentation approach that fully utilizes the large dataset and high-capacity neural networks to improve the $n$-gram LMs. Inspired by the recent advances in language model pre-training approaches~\cite{radford2018improving,devlin2019bert,radford2019language}, where the key idea is to pre-train the model on a unlabeled corpora and then fine-tune on different supervised downstream tasks, we introduce a general domain pre-training followed by in-domain fine-tuning strategy to construct high-capacity neural LMs for text generation. Specifically, a deep Transformer based neural LM is first pre-trained on a large general domain corpora, and then fine-tuned on the in-domain dataset. The well trained neural LM is then used to generate a large amount of high quality text to construct a synthetic corpus. The new $n$-gram LMs trained on the synthetic datasets are interpolated with baseline $n$-gram LMs and used for ASR decoding. 

This paper has two main contributions: (1) We are the first to leverage deep Transformer for text generation to improve ASR systems. In contrast to previous methods using shallow feedforward or recurrent neural networks~\cite{suzuki2019improvements}, our choice of deep Transformer network, which has been shown to excel at capturing long-term dependencies in text~\cite{vaswani2017attention,radford2018improving,devlin2019bert}, results in stronger modeling ability and therefore guarantees better generation quality. (2) We propose a general domain pre-training followed by in-domain fine-tuning strategy, which takes full advantage of the combination of large dataset and high-capacity models. Previous approaches with standard in-domain training are likely to encounter difficulties when there is very limited in-domain training data, either in optimization that leads to sub-optimal performances, or in generalization that results in text generated with poor diversity. The pre-training strategy contributes to overcome such limitations, making our approach more robust and generally applicable to different practical scenarios.

Experiments on two datasets show that our approach can effectively improve recognition accuracy over the strong baseline ASR systems and the existing LSTM based neural LM data augmentation methods. In particular, our proposed approach brings significant relative word error rate reduction up to $6.0\%$ for domains with limited in-domain data, which shows that our approach is very effective to improve speech recognition systems on new domains without extra efforts to collect in-domain data manually.

\section{Related Work}
\label{sec:related_work}
{\bf Neural LMs} Neural language models are proposed to overcome the curse of dimensionality by learning distributed word representations and probability function of word sequences~\cite{bengio2003neural}. Different neural network architectures have been proposed, including feedforward NN~\cite{bengio2003neural}, RNN~\cite{mikolov2010recurrent, mikolov2011extensions}, LSTM~\cite{sundermeyer2012lstm} and Transformer~\cite{vaswani2017attention,radford2018improving}, among which the self-attention based Transformer is the state-of-the-art architecture for many sequence modeling tasks due to its superiority in capturing longer-range linguistic structure. Although still less used in the first-pass ASR decoding due to its high inference latency, neural LMs have shown effectiveness in many other applications such as natural language generation~\cite{sutskever2011generating,masumura2015combinations}.

\noindent {\bf LM pre-training} Recent emerging language model pre-training approaches, such as OpenAI GPT~\cite{radford2018improving}, GPT2~\cite{radford2019language} and BERT~\cite{devlin2019bert}, have shown effectiveness for improving many natural language processing tasks. The key idea is to pre-train a deep Transformer network on unlabeled corpora then fine-tune the parameters on the downstream tasks. These approaches, however, mainly focus on the semi-supervised learning paradigm that leverages unsupervised pre-training to improve the supervised downstream tasks. In contrast, we focus mainly on the language modeling task and leverage the general domain pre-training to promote in-domain modeling. 

\noindent {\bf Improving $n$-gram LMs} Different methods have been proposed to improve $n$-gram LMs with neural LMs, including converting a feedforward nerual LM into an $n$-gram LM by directly assigning the probabilities ~\cite{adel2014comparing},  converting recurrent neural network (RNN) LM into backoff LMs and further improved quality with an iterative approach~\cite{arisoy2013converting}. The closest line of work to ours leverages multiple RNNLMs from different domains to generate text data for improving $n$-gram LMs~\cite{suzuki2019improvements}. However, their use of shallow RNN models and in-domain training restricts the generation ability. In contrast, our choice of pre-trained deep Transformer is able to generate text in higher quality by capturing longer range dependency, and better diversity through better model generalization.

\section{Approach}
\label{sec:approach}
We introduce the details of the proposed data augmentation method for improving $n$-gram LMs in this section. 

\subsection{Overall Pipeline}
\begin{figure}[t]
    \centering
    \includegraphics[width=0.9\linewidth]{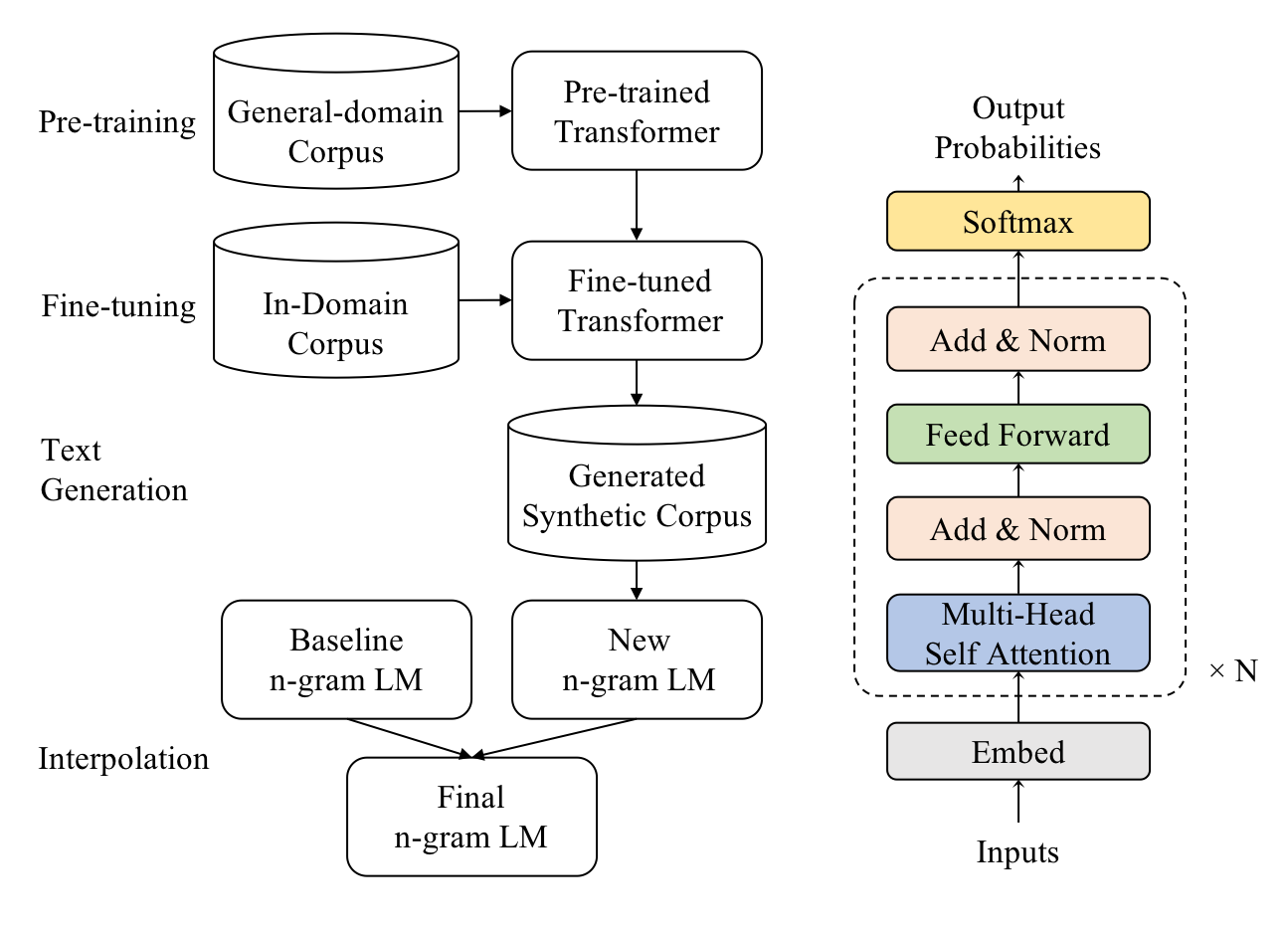}
    \caption{ {\bf (left)} Overall pipeline of the proposed data augmentation approach. {\bf (right)} Transformer architecture for neural LM.}
    \label{fig:pipeline}
\end{figure}

The overall pipeline of the proposed approach is depicted in Fig~\ref{fig:pipeline} (left), which consists of four steps including pre-training, fine-tuning, generation and interpolation. Specifically, we first pre-train a deep Transformer on the large general-domain corpus and then fine-tune on the target domain dataset. We use the obtained neural LM to generate large amount of high quality in-domain text data, which is then used to construct a synthetic dataset for $n$-gram LM training. The new $n$-gram LMs are eventually interpolated with the previous baseline $n$-gram LMs and evaluated in the ASR system.

\subsection{Pre-training and Fine-tuning}

We propose a general-domain pre-training followed by in-domain fine-tuning strategy. Given a large and diverse collection $\mathcal{C}=\{X_1, \dots, X_N\}$, where each $X_i$ is a sequence of word or subword units $X_i=\{x^{(i)}_1, \dots, x^{(i)}_k\}$, we use the standard left-to-right language modeling objective to maximize the likelihood:

\begin{equation}
    \mathcal{L}(\mathcal{C}) = \sum_i \sum_t logP(x^{(i)}_t|x^{(i)}_{<t}; \Theta)
\label{eqn:lm}
\end{equation}
where the conditional probability $P$ is modeled by a neural network with parameters $\Theta$ and $x^{(i)}_{<t}$ is the history up to $t$.

Fine-tuning is then performed for each domain of interest. Given a in-domain collection $\mathcal{D}=\{Y_1, \dots, Y_M\}$, $Y_i=\{y^{(i)}_1, \dots, y^{(i)}_k\}$, the model is trained using the same objective in Eqn.~\ref{eqn:lm}. The model parameters $\Theta$ are initialized by the pre-trained model and further optimized on $\mathcal{D}$ till converge. 

In this work, we use the deep Transformer decoder~\cite{radford2018improving}, a variant of Transformer~\cite{vaswani2017attention}, as the architecture for our neural LM $\Theta$. As is illustrated in Fig~\ref{fig:pipeline} (right), the model is composed of a stack of $N$ transformer blocks. Each block has two types of basic layers: (1) Multi-head self attention layer, which generates an adaptive weighted sum of the input hidden representations from previous layer; (2) Feed forward layer, which applies non-linear transformation to the hidden vector. Each basic layer is associated with layer normalization, residual connection and dropout. 

This is the first work that leverages deep Transformer for text generation for ASR. Intuitively, deep Transformer is superior to previous shallow feed forward or recurrent neural LMs in two ways: (1) the self attention mechanism eases the challenge of long-range dependency learning, which is particularly important for high-quality text generation; (2) the high model capacity and depth leads to better modeling and generalization ability. Our proposed pre-training strategy helps overcome previous limitations in lack of in-domain training data by making use of the largely available general-domain data, and makes it possible to construct such strong neural LM.

\subsection{Text Generation}
We generate a large amount of text data with the well trained neural LM. The generation is performed by sampling from the model distribution given the prefixed context. Specially, we first construct a prefix corpus $\mathcal{D}_P=\{S_1, \dots, S_n\}$, where $S_i=\{s^{(i)}_1, \dots, s^{(i)}_k\}$ with $k$ prefix tokens. During text generation, $S_i$ is fed into the model as the prefixed context. The neural LM produces probabilities over the vocabulary $\mathcal{V}$:

\begin{equation}
    p_i = \frac{exp(\frac{h_i}{\tau})}{\sum_j exp(\frac{h_j}{\tau})}
\end{equation}
where $h=(h_1, \dots, h_{|\mathcal{V}|})$ is the logit vector, and $\tau$ is the temperature for sampling. Then the output tokens are sampled from the probability distribution. 

A rule-based data filtration is performed on the generated synthetic corpus to ensure data quality for $n$-gram LM training. We designed multiple different rules including filter by maxmimum and minimum sequence length, filter by out-of-vocabulary words, filter by domain-specific keyword, and filter by number of duplicated generation. The thresholds for each filtering rule are selected based on data distribution of the in-domain training data. 

\section{Experiment}
\label{sec:exp}
We evaluate the effectiveness of the proposed approach on two speech recognition datasets. We compare the performance of deep Transformer with shallow LSTMs, as well as the pre-training strategy with the traditional in-domain training. As we will show, the proposed combination of pre-training strategy and deep Transformer lead to substantial improvements.

\subsection{Datasets}
We conduct experiments on two in-house speech recognition datasets: the speech assistant dataset (denoted as {\em Assistant}), and the conversational speech dataset (denoted as {\em Conversation}). Both datasets are completely anonymized and no user-identifiable information (UII) is access to both annotators and researchers. 

\noindent {\bf Assistant} The Assistant dataset consists of English utterances that are commands users give to Facebook Portal\footnote{Portal is a video enabled smart device.} after the wakeword ``Hey Portal" to carry out certain actions. The utterances can be categorized into various sub-domains by the type of actions, such as making phone calls to their friends ({\em calling}),  and device control ({\em device}), or getting weather information ({\em weather}). We use a mixed set of utterances that are randomly sampled from both internal dogfooding and Facebook Portal live traffic. Internal dogfooding is an activity from internal employees with signed agreements to have their activity reviewed and tested. We choose to exclude some domains that contain limited utterance patterns such as {\em calling} as enriching the training data is not helpful for these domains. All these sampled utterances are voice morphed before sending to annotators for transcription. In total, we use a collection of $730$k utterances as training data, $15$k as development data, and $40$k as test data. 

\noindent {\bf Conversation} The Conversation dataset was collected through crowd-sourcing. It consists of conversations between each pair of crowd-sourcers with more than $20$ topics that are commonly mentioned in daily life, including family, travel, etc. We split the data into training ($240$k), development ($7$k), and test ($20$k) sets. 

\noindent {\bf General-domain Pre-training} We use a large in-house English text corpus as general domain data for neural LM pre-training, which contains a random sample of $110$M public posts and comments users shared on Facebook. We use byte pair encoding
(BPE\footnote{https://github.com/glample/fastBPE})~\cite{sennrich2016neural} to segment word
tokens into subword units, forming a $25k$-subword vocabulary used for both Assistant and Conversation dataset. We directly converted the text data into machine reading format for model training and did not manually look into the actual content. 

\subsection{Experiment Setups}
\noindent {\bf Baselines} We compare our proposed approaches with two baselines, including (1) baseline $n$-gram without augmentation ({\em Baseline})~\cite{duc2019asr}, and (2) data augmentation with text generated by LSTM trained on in-domain data~\cite{suzuki2019improvements} ({\em LSTM (in-domain)}). 

\noindent {\bf ASR System} We use a state-of-the-art hybrid ASR system that utilizes multi-layer Latency Controlled Bidirectional Long Short-Term Memory RNNs (LC-BLSTM) for acoustic modeling with grapheme representations. And it uses pruned 4-gram LMs in the first-pass decoding with an in-house dynamic decoder, where the final LMs are interpolated with LMs trained from both in-domain and general-domain training data. For each approach on each dataset, we optimize all model hyper-parameters on the development sets.

\noindent {\bf Model Settings} We adopt the \texttt{GPT} configuration following~\cite{radford2018improving}, with the dimension of word embeddings, hidden states and non-linear layers set as $768$, $768$ and $3072$ respectively. The numbers of both decoder blocks and attention heads are set as $12$, and the dropout rate is $0.1$. For the LSTM baseline, we adopt a model with similar model size as Transformer for fair comparison. We use a stack of $2$ LSTM layers, where the dimension of word embeddings, hidden states set as $1024$ and $2048$ respectively. The dropout rate is $0.2$. We use the Adam optimization scheme following~\cite{radford2018improving}. The models are trained on $16$ V100 GPUs, and based on the PyTorch implementation of Transformer\footnote{https://github.com/pytorch/fairseq}.

\noindent {\bf Text Generation} We extract the prefix sequences with $k$ tokens from the in-domain training data, where $k\in\{1,2,3,4,5,6\}$. We keep the top $5$ sampling hypotheses with length penalty set as $1.0$. The temperature for sampling is set as $\tau \in \{1.0, 1.1,\dots, 1.5\}$ for the best balance of generation quality and diversity. 

\noindent {\bf Evaluation} For evaluation, we interpolate the new $n$-gram LM with the baseline $n$-grams and evaluate the methods via word error rate (WER) of the ASR system, and report WER reductions (WERR) over the baseline approach.

\subsection{Assistant}

\begin{table}[t]
\caption{The overall relative word error rate reduction (WERR) for each data augmentation approach on Assistant. }
\centering
\begin{tabular}{lcc}
\toprule
                          & WERR           \\ 
\midrule
LSTM (in-domain)          & $1.07\%$       \\
Transformer (in-domain)   & $1.30\%$       \\
Transformer (pre-trained) & ${\bf 2.25\%}$ \\ 
\bottomrule
\end{tabular}
\label{tbl:portal_overall}
\end{table}

\begin{table}[]
\caption{Relative word error rate reduction (WERR) for each data augmentation approach on different Assistant sub-domains, including device, weather and music.}
\centering
\begin{tabular}{lccc}
\toprule
                          & Device         & Weather        & Music          \\
\midrule
LSTM (in-domain)          & $3.16\%$       & $1.64\%$       & $2.42\%$       \\
Transformer (in-domain)   & $3.56\%$       & $3.10\%$       & $3.52\%$       \\
Transformer (pre-trained) & ${\bf 3.75\%}$ & ${\bf 6.01\%}$ & ${\bf 4.91\%}$ \\
\bottomrule
\end{tabular}
\label{tbl:portal_subdomain}
\end{table}

\begin{table}[t]
\caption{Word-level perplexity of neural LMs on Assistant test set.}
\centering
\begin{tabular}{lcc}
\toprule
                          & Pre-trained & Fine-tuned   \\
\midrule
LSTM (in-domain)          & $-$         & $7.56$       \\
Transformer (in-domain)   & $-$         & $7.11$       \\
Transformer (pre-trained) & $55.96$     & $\bf{6.25}$  \\
\bottomrule
\end{tabular}
\label{tbl:portal_ppl}
\end{table}

\begin{table}[]
\centering
\small
\caption{Examples generated by Transformer pre-trained and fine-tuned on Assistant in-domain data. The examples are excluded from the in-domain training data.
}
\begin{tabular}{ll}
\toprule
Domain  & Examples                                       \\
\midrule
        & replay the current track                       \\
Music   & what album is this track from                  \\
        & play french playlist on spotify                \\
\midrule
        & what's the hourly forecast for today           \\
Weather & what's the weather in youngstown ohio          \\
        & what's the temperature in delray beach florida \\
\bottomrule
\end{tabular}
\label{tbl:portal_cases}
\end{table}

We report the overall word error rate reduction over the baseline approach on Assistant and several sub-domains including {\em device}, {\em weather} and {\em calling} in Table~\ref{tbl:portal_overall} and Table~\ref{tbl:portal_subdomain}, respectively. From these tables we have the following observations: 

1. The proposed data augmentation approaches effectively improve the overall quality of $n$-gram LMs in the ASR systems. In particular, our pre-trained deep Transformer achieves over $2.2\%$ relative reduction in WER, which significantly outperforms the LSTM-based approach. 

2. The proposed approach is particularly beneficial for sub-domains with less training data. Due to the unbalance of Assistant dataset, some sub-domains like {\em weather} and {\em music} are important yet have only a small training collection. The proposed approach with pre-trained Transformer achieves over $6\%$ and $4.9\%$ relative WER reduction in the {\em weather} and {\em music} sub-domains and outperforms augmentation with LSTM or Transformer constructed with traditional training strategy without pre-training by large margin. For the large domains such as {\em device}, we observe fewer gains from pre-training as the in-domain data is already sufficient to train a neural LM with good performance in these large domains. However, we can still see that Transformer outperforms LSTM, and pre-training slightly further improves the performance. 

3. The improvements have been brought by both use of deep Transformer architecture and pre-training strategy. WER reduction is observed by replacing the LSTM to deep Transformer network for neural LM, which indicates the superiority of the model architecture. With the pre-training strategy, the model can be even better utilized and results in the best ASR decoding performance.

We further present detailed analysis on the different neural LMs. Table~\ref{tbl:portal_ppl} shows the perplexity of neural LMs with different architecture and training strategy, which verifies that deep Transformer has better modeling performance than the previous LSTM/RNN, and demonstrates that the general domain pre-training and in-domain fine-tuning strategy is an important component for high-quality deep model construction. Table~\ref{tbl:portal_cases} presents multiple generated cases in the {\em music} and {\em weather} sub-domains, which are not originally included in the in-domain Assistant training collection. The examples illustrate that the pre-trained neural LM can generate high-quality text for data augmentation and enrich the sequence patterns to help ease the problem of data sparsity for $n$-gram LMs.

\subsection{Conversation}

\begin{table}[t]
\caption{Relative word error rate reduction (WERR) over the baseline approach on Conversation test set. ``\#Aug" denotes the number of augmented training data.}
\centering
\begin{tabular}{lcc}
\toprule
                          & \#Aug   & WERR          \\ 
\midrule
LSTM (in-domain)          & $4$M    & $1.00\%$      \\
Transformer (in-domain)   & $4$M    & $1.86\%$      \\
Transformer (pre-trained) & $4$M    & $\bf{2.62\%}$ \\ 
\midrule
LSTM (in-domain)          & $18$M   & $1.66\%$      \\
Transformer (in-domain)   & $18$M   & $2.12\%$      \\
Transformer (pre-trained) & $18$M   & $\bf{3.08\%}$ \\ 
\midrule
\end{tabular}
\vspace{-5pt}
\label{tbl:ocean_overall}
\end{table}

\begin{table}[t]
\caption{Word-level perplexity of neural LMs on Conversation test set. ``pre-trained" denotes model pre-trained on general background data and then fine-tuned on Conversation dataset. ``in-domain" denotes model trained only on Conversation.}
\centering
\begin{tabular}{lcc}
\toprule
                          & Pre-trained & Fine-tuned \\
\midrule
LSTM (in-domain)          & $-$         & $112.45$   \\
Transformer (in-domain)   & $-$         & $89.36$    \\
Transformer (pre-trained) & $80.43$     & $46.68$    \\
\bottomrule
\end{tabular}
\vspace{-5pt}
\label{tbl:ocean_ppl}
\end{table}

We further evaluate the approach on the Conversation dataset, which has a much smaller training collection ($240$k) than Assistant ($730$k), with more complex and diverse patterns. As can been seen from Table~\ref{tbl:ocean_ppl}, the pre-trained deep Transformer demonstrates significant superiority over the neural LMs with traditional training scheme in such a scenario with the lack of in-domain training data.

The performances are presented in Table~\ref{tbl:ocean_overall}. With $4$ million instances of synthetic in-domain training corpus, the proposed approach with pre-trained deep Transformer achieves over $2.6\%$ relative WER reduction, compared with $1.8\%$ relative reduction of Transformer and $1.0\%$ of LSTM with traditional in-domain training strategy. The performance continues to grow when we enlarge the volume of generated data to $18$ millions, and achieves over $3.0\%$ relative WER reduction over the strong baseline system. 

These results corroborate our motivation and demonstrate that:

1. The proposed approach is simple yet effective in improving $n$-gram LMs in ASR. The number of augmented training data can be easily scaled up for further decoding performance improvements with minimal computational cost.

2. The general domain pre-training then in-domain fine-tuning strategy is the key component of the proposed method. The superiority of pre-training over the traditional in-domain training strategy is at two scales: (i) The large-scale general pre-training enables construction of the state-of-the-art deep Transformer rather than shallow RNNs~\cite{suzuki2019improvements}, which leads to strong neural LMs with large model capacity and better generalization to generate text with both high quality and good diversity. (ii) In the cases with lack of in-domain training data, direct in-domain training results in sub-optimal performances of the neural LMs (Table~\ref{tbl:ocean_ppl}). The pre-training stategy overcomes the problem, making it more robust and generally applicable to different scenarios.

\section{Conclusion}
\label{sec:conclusion}
In this paper, we introduce a text generation based data augmentation approach that effectively improves $n$-gram LMs and achieves better ASR decoding accuracy. Our contributions are at two scales: (1) We are the first to leverage deep Transformer for text generation for ASR systems; (2) We proposed a general domain pre-training followed by in-domain fine-tuning strategy that enables us to fully leverage the large corpora and the high-capacity neural networks. The approach is general and widely applicable to different data domains to help improve the first-pass decoding accuracy of the ASR systems. 


\bibliographystyle{IEEEbib}
\bibliography{mybib}

\end{document}